%% file: main.tex
\documentclass[10pt,twocolumn,letterpaper]{article}

\usepackage[pagenumbers]{iccv} % To force page numbers, e.g. for an arXiv version


\usepackage{xcolor}         % colors
\usepackage{listings}
\usepackage{wrapfig}
\usepackage{times}
\usepackage{epsfig}
\usepackage{graphicx}
\usepackage{amsmath}
\usepackage{amssymb}
\usepackage{bm}
\usepackage{multirow}
\usepackage{bbding}
\usepackage[linesnumbered,ruled,vlined]{algorithm2e}

\usepackage{caption}    

\usepackage{diagbox}
\usepackage{booktabs}
\usepackage{dsfont}
\usepackage{color}  
\newcommand{\tabincell}[2]{\begin{tabular}{@{}#1@{}}#2\end{tabular}} 

\usepackage{bm}
\usepackage{multirow}
\usepackage{bbding}

\definecolor{iccvblue}{rgb}{0.21,0.49,0.74}
\usepackage[pagebackref,breaklinks,colorlinks,allcolors=iccvblue]{hyperref}

\def\paperID{2636} % *** Enter the Paper ID here
\def\confName{ICCV}
\def\confYear{2025}

\title{Semantic Alignment and Reinforcement for Data-Free Quantization of Vision Transformers}

\author{
Yunshan Zhong$^{1,2}$, Yuyao Zhou$^2$, Yuxin Zhang$^2$, Wanchen Sui$^3$, Shen Li$^3$,  \\  Yong Li$^3$,Fei Chao$^2$, Rongrong Ji$^{1,2}$\thanks{Corresponding Author: rrji@xmu.edu.cn}\\
$^1$Institute of Artificial Intelligence, Xiamen University\\
$^2$MAC Lab, Department of Artificial Intelligence, School of Informatics, Xiamen University \, \\ $^3$Alibaba  \\
{\tt\small zhongyunshan@stu.xmu.edu.cn, yuyaozhou@stu.xmu.edu.cn} \\ 
{\tt\small  yuxinzhang@stu.xmu.edu.cn, wanchen.swc@alibaba-inc.com}\\ {\tt\small litan.ls@alibaba-inc.com, jiufeng.ly@alibaba-inc.com, fchao@xmu.edu.cn, rrji@xmu.edu.cn}
}

\begin{document}
\maketitle
\input{sec/0_abstract}

{
    \small
    \bibliographystyle{ieeenat_fullname}
    \bibliography{main}
}

\end{document}

%% file: sec/0_abstract.tex
\begin{abstract}

Data-free quantization (DFQ) enables model quantization without accessing real data, addressing concerns regarding data security and privacy. With the growing adoption of Vision Transformers (ViTs), DFQ for ViTs has garnered significant attention. 
%
% However, existing DFQ methods suffer from two critical limitations: (1) semantic distortion, in which the semantics of synthetic images deviate substantially from those of real images, and (2) semantic inadequacy, where synthetic images contain large regions with minimal content and overly simplistic textures, ultimately resulting in suboptimal quantization performance. 
% %
However, existing DFQ methods exhibit two limitations: (1) semantic distortion, where the semantics of synthetic images deviate substantially from those of real images, and (2) semantic inadequacy, where synthetic images contain extensive regions with limited content and oversimplified textures, leading to suboptimal quantization performance.
To address these limitations, we propose SARDFQ, a novel \textbf{S}emantics \textbf{A}lignment and \textbf{R}einforcement \textbf{D}ata-\textbf{F}ree \textbf{Q}uantization method for ViTs. To address semantic distortion, SARDFQ incorporates Attention Priors Alignment (APA), which optimizes synthetic images to follow randomly generated structure attention priors.
To mitigate semantic inadequacy, SARDFQ introduces Multi-Semantic Reinforcement (MSR), leveraging localized patch optimization to enhance semantic richness across synthetic images.
Furthermore, SARDFQ employs Soft-Label Learning (SL), wherein multiple semantic targets are adapted to facilitate the learning of multi-semantic images augmented by MSR.
Extensive experiments demonstrate the effectiveness of SARDFQ, significantly surpassing existing methods. For example, SARDFQ improves top-1 accuracy on ImageNet by 15.52\% for W4A4 ViT-B\footnote{The code  is at \url{https://github.com/zysxmu/SARDFQ}.}.

\end{abstract}

\begin{figure}[htbp]
\centering
\begin{subfigure}{\linewidth}
    \centering
    \includegraphics[width=0.75\linewidth]{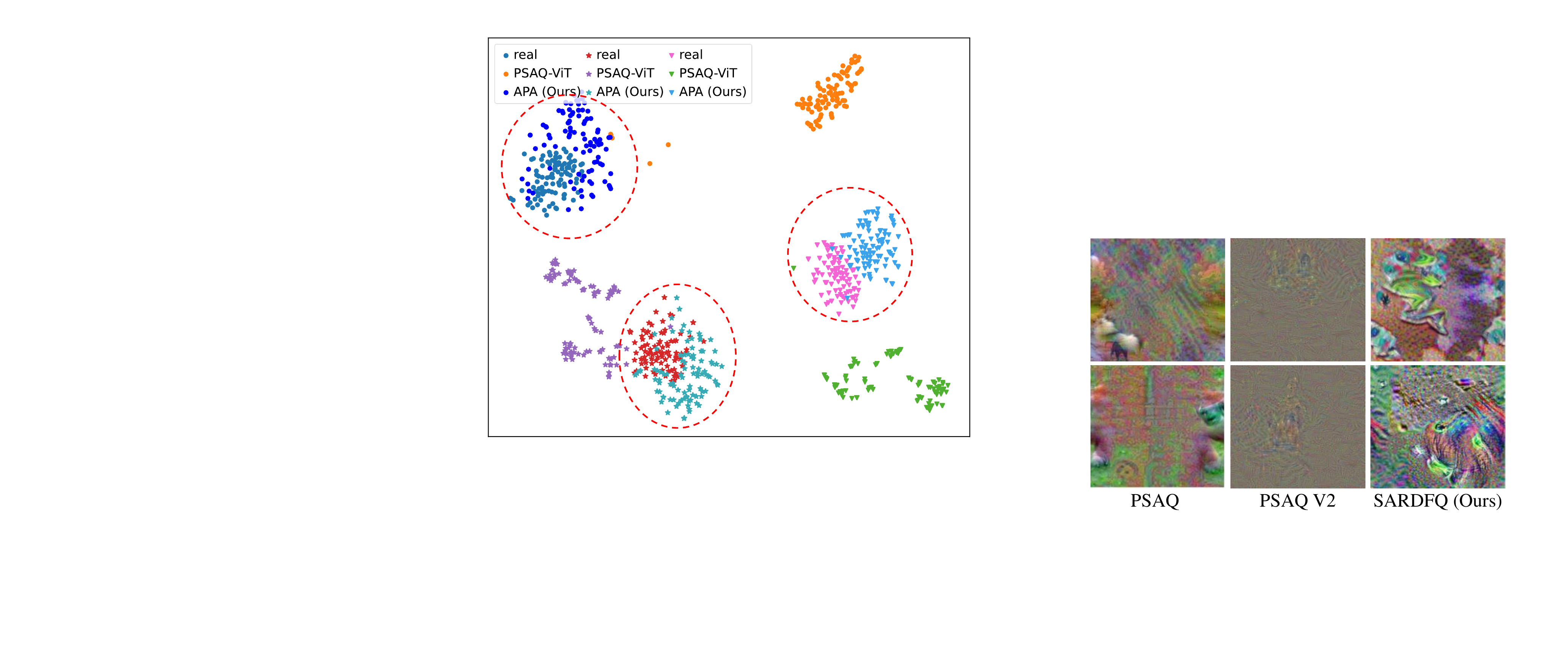}
    \caption{
    % Visualization of penultimate features of the synthetic images using t-SNE~\cite{tSNE}. Each shape (circle, star, triangle) represents a category. The red dashed circles highlight the features of using our APA and real images. The features are extracted from the penultimate layer of DeiT-S. The features of PSAQ-ViT~\cite{li2022patch} deviate significantly from those of the real images, indicating semantic distortion. While applying our APA leads to a closer alignment with real images, indicating aligned semantics.
    %
    The t-SNE\cite{tSNE} visualization of the penultimate-layer features (extracted by DeiT-S) of synthetic images. Each marker (circle, star, triangle) represents a distinct category. The red dashed circles highlight the features extracted from our APA and real images. Notably, the features produced by PSAQ-ViT\cite{li2022patch} exhibit substantial deviation from those of real images, indicating semantic distortion. In contrast, our APA yields features that more closely align with those of real images, suggesting improved semantics alignment.
    }
    \label{fig:insight-tsne}
\end{subfigure}

\begin{subfigure}{\linewidth}
\centering
        \includegraphics[width=0.75\linewidth]{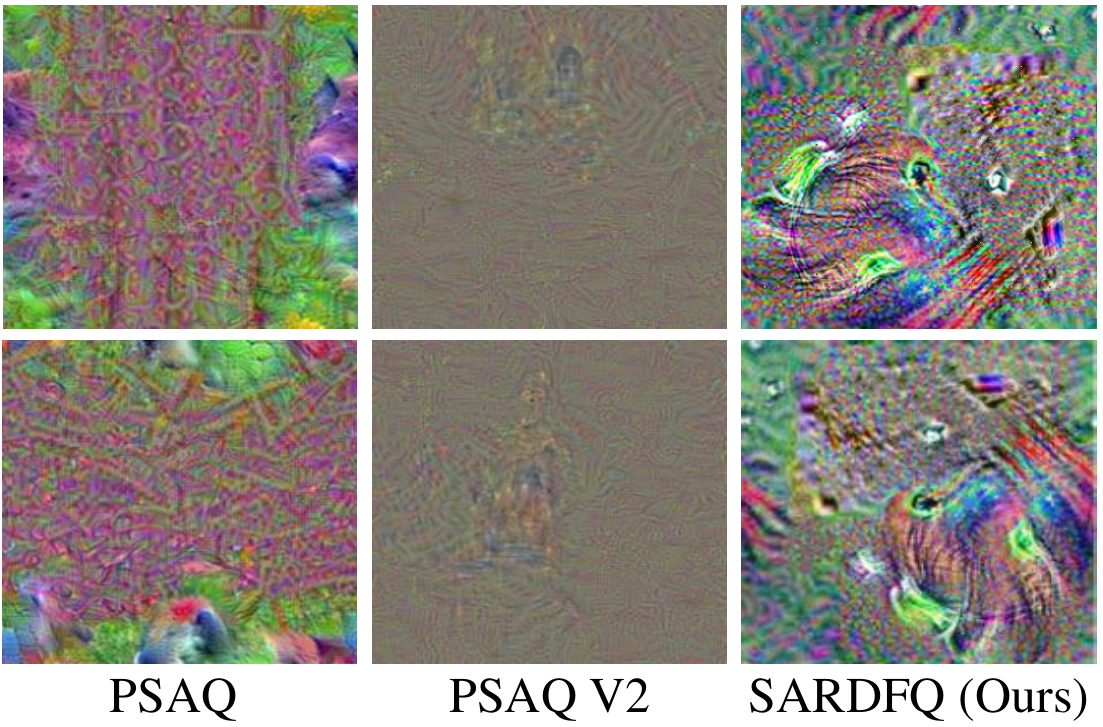}
        \caption{
        % The images of PSAQ-ViT and PSAQ-ViT V2 contain many dull regions that show limited content and simplified textures, indicating semantic inadequacy. Our SARDFQ shows greater diversity in content and texture overall, indicating reinforced semantics.
        %
        The images of PSAQ-ViT and PSAQ-ViT V2 exhibit numerous dull regions with limited content and simplified textures, reflecting semantic inadequacy. In comparison, our SARDFQ generates images with greater diversity in both content and texture, demonstrating enhanced semantics.
        } 
        \label{fig:insight-inadequacy}
\end{subfigure}
\caption{Illustration of (a) semantic distortion and (b) semantic inadequacy.}
\label{fig:insights}
\vspace{-3mm}
\end{figure}

% \begin{figure}
% \centering
% \includegraphics[width=0.95\linewidth]{t-SNE.pdf} 
% \caption{
% Feature visualization using $t$-SNE~\cite{tSNE}. Each shape (circle, star, triangle) represents a category. The red dashed circles highlight the features of our APA and real images. The visualized features are activations extracted before the classification head of DeiT-S.
% %
% The features from PSAQ-ViT~\cite{li2022patch} deviate significantly from those of the real images, suggesting semantic distortion. While the features from our APA are more closely aligned with those of real images, suggesting aligned semantics.
% }
% \label{fig:insight-tsne}
% \end{figure}

\section{Introduction}

% Vision transformers (ViTs)~\cite{dosovitskiyimage} have attracted widespread interest from both academia and industry~\cite{khan2021transformers,han2020survey} due to their superior performance across various vision tasks~\cite{touvron2021training,carion2020end,arnab2021vivit,ma2023towards}.
% %
% However, the substantial computational demands and high memory requirements of ViTs pose significant challenges for deployment in resource-limited environments~\cite{tang2022patch,jia2021efficient,hou2022multi,zheng2023less,li2023vit,liu2021post,chen2023diffrate}.
% %
% To address this challenge, model quantization~\cite{whitepaper} has emerged as a promising solution, reducing model complexity by enabling a low-bit representation of weights and activations.
% %
% Nevertheless, most existing quantization methods rely on access to the original training data, raising concerns regarding data privacy and security. This data dependency restricts their application in data-sensitive environments~\cite{cai2020zeroq,zhong2022intraq,guo2022squant,yin2020dreaming}.
% where such access is prohibited

Vision Transformers (ViTs)~\cite{dosovitskiyimage,khan2021transformers,han2020survey} have demonstrated remarkable success across various computer vision tasks~\cite{touvron2021training,carion2020end,arnab2021vivit,ma2023towards}. However, their high computational cost and substantial memory footprint hinder deployment in resource-constrained environments~\cite{tang2022patch,jia2021efficient,hou2022multi,zheng2023less,li2023vit,liu2021post,chen2023diffrate}. To address these limitations, quantization~\cite{whitepaper} has emerged as a promising solution, which reduces model complexity by converting full-precision weight and activations into low-bit representations.

% Traditional quantization methods rely on access to the original training dataset, raising data privacy and security concerns in real-world applications~\cite{cai2020zeroq,zhong2022intraq,guo2022squant,yin2020dreaming}.
Traditional quantization methods typically require access to the original training dataset, which raises data privacy and security concerns~\cite{cai2020zeroq, zhong2022intraq, guo2022squant, yin2020dreaming}.
%
% Consequently, data-free quantization (DFQ) has gained increasing attention, enabling quantization without requiring real data~\cite{xu2020generative,zhang2021diversifying,choi2021qimera}.
As a result, data-free quantization (DFQ) has gained increasing attention, allowing quantization without the need for real data~\cite{xu2020generative,zhang2021diversifying,choi2021qimera}.
%
% Most existing DFQ methods are tailored to convolutional neural networks (CNNs) and are not directly applicable to ViTs.
% %
% These methods typically utilize batch normalization statistics (BNS), which capture the distribution of real data, to synthesize in-distribution synthetic data~\cite{cai2020zeroq,xu2020generative,zhang2021diversifying,zhong2022intraq}.
%
However, most existing DFQ methods are designed specifically for convolutional neural networks (CNNs) and are not directly applicable to vision transformers (ViTs). These methods generally rely on batch normalization statistics (BNS), which capture the distribution of real data, to synthesize in-distribution synthetic data~\cite{cai2020zeroq, xu2020generative, zhang2021diversifying, zhong2022intraq}.
%
%
% However, BNS is unavailable for ViTs that use layer normalization (LN) to dynamically compute distribution statistics during inference~\cite{li2022patch}.
Yet, BNS is unavailable for ViTs, which use layer normalization (LN) to dynamically compute distribution statistics during inference~\cite{li2022patch}.
%
% Recently, several works have been proposed to accommodate the unique structure of ViTs~\cite{li2022patch,li2023psaqv2,ramachandran2024clamp,choi2024mimiq,husparse}. For example, PSAQ-ViT~\cite{li2022patch} introduces patch similarity entropy (PSE) loss to optimize Gaussian noise towards usable synthetic images.
%
Recently, several DFQ methods have been proposed for ViTs~\cite{li2022patch, li2023psaqv2, ramachandran2024clamp, choi2024mimiq, husparse}. For example, PSAQ-ViT~\cite{li2022patch} introduces patch similarity entropy (PSE) loss to optimize Gaussian noise towards usable synthetic images.

Nevertheless, we observe that existing methods suffer from semantic distortion and inadequacy issues. 
As shown in Fig.\,\ref{fig:insight-tsne}\footnote{We do not plot the results of images from PSAQ-ViT V2 as they lack specific categories.}, features of synthetic images generated by PSAQ-ViT deviate significantly from those of real images. Tab.\,\ref{tab:insight} shows that the cosine similarity between synthetic images from PSAQ-ViT and real images is notably low, also indicating significant distortion. These results highlight the issue of \textbf{semantic distortion}.
Moreover, as shown in Fig.\,\ref{fig:insight-inadequacy}\footnote{More results are presented in the appendix.}, synthetic images generated by PSAQ-ViT and PSAQ-ViT V2 exhibit many regions with limited content diversity and overly simplified textures. These low-quality dull regions are useless or even detrimental to model learning~\cite{husparse}, highlighting the issue of \textbf{semantic inadequacy}.
Consequently, quantized models trained on such low-quality images suffer from degraded performance.

Motivated by the above analysis, we propose a novel \textbf{S}emantics \textbf{A}lignment and \textbf{R}einforcement \textbf{D}ata-\textbf{F}ree \textbf{Q}uantization method for ViTs, termed SARDFQ. The overall framework is depicted in Fig.\,\ref{fig:framework}. 
To address the semantic distortion issue, SARDFQ introduces Attention Priors Alignment (APA), where synthetic images are optimized to follow structured attention priors generated using Gaussian Mixture Models (GMMs). APA effectively aligns the semantics of synthetic images with real images, as validated by both visual and quantitative analyses. As shown in Fig.\,\ref{fig:insight-tsne}, features of APA exhibit a closer alignment to those of real images, while quantitative results in Tab.\,\ref{tab:insight} also confirm that the semantics of APA are more consistent to the real images, indicating enhanced semantics alignment.
To address the semantic inadequacy issue, SARDFQ incorporates Multi-Semantic Reinforcement (MSR) and Softlabel Learning (SL). MSR utilizes localized patch optimization, which encourages different sub-patches of synthetic images to capture various semantics, reinforcing the rich semantics across images. SL applies multiple semantic targets to accommodate the learning of multi-semantic images augmented by MSR. As shown in Fig.\,\ref{fig:insight-inadequacy}, synthetic images after applying MSR exhibit greater diversity in content and texture, providing reinforced semantics.

Experimental results across various ViT models and tasks demonstrate that SARDFQ presents substantial performance improvements. For example, SARDFQ achieves a 15.52\% increase in top-1 accuracy on the ImageNet dataset for the W4A4 ViT-B model.

\section{Related Works}

\subsection{Vision Transformers}

The great success of transformers in the natural language processing field has driven widespread attempts in the computer vision community to apply them to vision tasks~\cite{chen2021transformer,wu2021rethinking,han2021transformer}.
%
% By treating images as a sequence of flattened 2D patch tokens,
ViT~\cite{dosovitskiyimage} is the pioneer that builds a transformer-based model to handle images, boosting the performance on the image classification task.
%
% To address the dependency on large datasets,
DeiT~\cite{touvron2021training} introduces an efficient teacher-student training strategy where a distillation token is employed to distill knowledge from the teacher model to the student model.
Swin Transformers~\cite{liu2021swin} builds an efficient and effective hierarchical model by introducing a shifted window-based self-attention mechanism.
% to perform local attention and thus has linear computational complexity to image size.
%
Other than the image classification task, the applications of ViTs also have broadened considerably, manifesting groundbreaking performance in object detection~\cite{carion2020end}, image segmentation~\cite{chen2021pre,zheng2021rethinking}, low-level vision~\cite{liang2021swinir}, video recognition~\cite{neimark2021video,arnab2021vivit}, and medical image processing~\cite{shamshad2023transformers}, \emph{etc}.
Nevertheless, the impressive performance of ViTs relies on a high number of parameters and significant computational overhead, preventing deployment in resource-constrained environments.
Several recent efforts design lightweight ViTs, such as MobileViT~\cite{mehta2021mobilevit}, MiniVit~\cite{zhang2022minivit}, and TinyViT \cite{wu2022tinyvit}. However, the model complexity is still unsatisfactory~\cite{li2022patch}.

\iffalse
\begin{figure}% 靠文字内容的左侧
\centering
\subfloat[PSAQ-ViT]{
\includegraphics[width=0.95\linewidth]{attn_vis-psaq.pdf} \label{fig:attn_vis-psaq}
}
\\
\subfloat[SARDFQ (Ours)]{
\includegraphics[width=0.95\linewidth]{attn_vis-sfp.pdf} \label{fig:attn_vis-sfp}
}
\\
\subfloat[Real Images]{
\includegraphics[width=0.95\linewidth]{attn_vis-real.pdf} \label{fig:attn_vis-real}
}
\caption{Heatmap of attention maps of the classification token within the block.11 in DeiT-S. The used images belong to the same category. The brighter the color, the higher the attention score. Each column represents a head.}
\label{fig:attn_vis}
\end{figure}
\fi

\begin{figure*}[!t]
\centering
\includegraphics[width=0.85\linewidth]{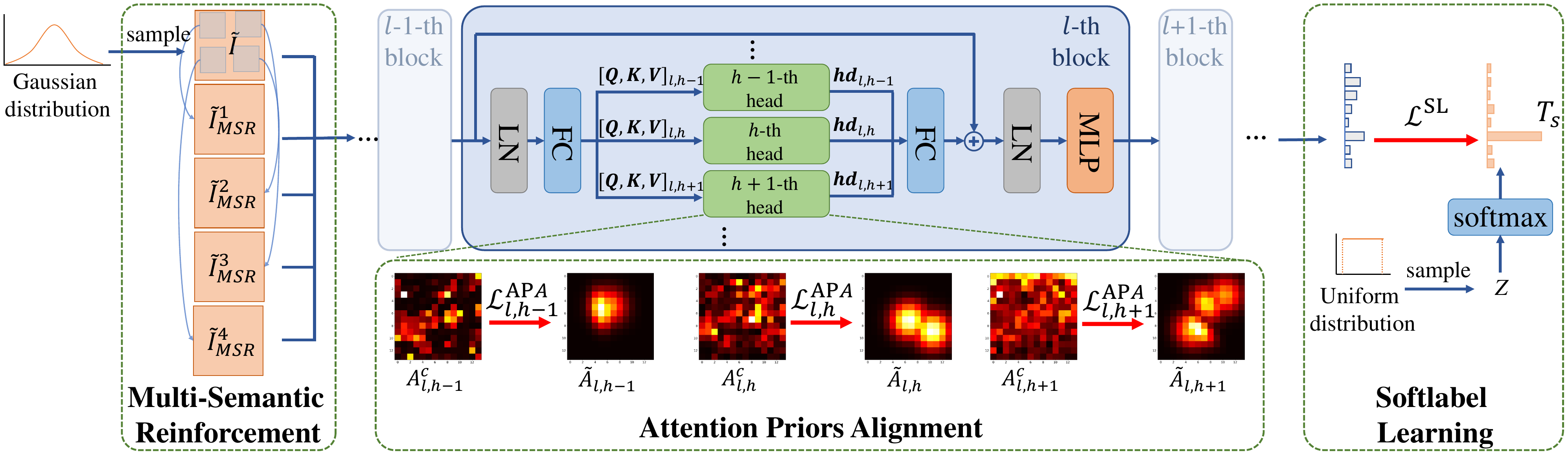}
\caption{SARDFQ Framework overview: 
Attention Priors Alignment (APA) employs randomly generated attention priors to improve semantics alignment.
Multi-Semantic Reinforcement (MSR) learns the different regions of synthetic images with various semantics to enhance overall semantic richness. Meanwhile, Softlabel Learning (SL) adopts multiple semantic targets to ensure consistent learning of multi-semantic images augmented by MSR.
} 

\label{fig:framework}
\vspace{-3mm}
\end{figure*}

\subsection{Network Quantization}

\textbf{Data-Driven Quantization.}
Model quantization reduces the complexity of neural networks by replacing full-precision weight and activation with the low-bit format.
Data-driven quantization can be roughly divided into two categories: quantization-aware training (QAT) and post-training quantization (PTQ).
QAT is compute-heavy since it re-trains the quantized model with the full training data to retain performance~\cite{li2022q,li2023vit,APoT,gong2019differentiable,Yang2023Oscillation,LSQ,huang2023variation,sun2024p4q}. 
PTQ perform quantization with a tiny dataset and a reduced time overhead, harvesting widespread attention~\cite{ACIQ,li2021brecq}.
The specific architecture of ViTs, such as LayerNorm and the self-attention module, urges distinct PTQ methods compared to CNNs~\cite{liu2023noisyquant,ding2022towards,lin2022fqvit,li2023vit,frumkin2023jumping,zhongerq}.
For example, Liu \emph{et al}.~\cite{liu2021post} develop a ranking loss to maintain the relative order of the self-attention activation.
%
% RepQ-ViT~\cite{li2023repq} deploys a complex quantizer for high variance activation at first and then simplifies the quantizer through re-parameterization.
%
Unfortunately, both QAT and PTQ involve the original training data, causing concerns about data privacy and security issues in data-sensitive scenarios.

\textbf{Data-Free Quantization.}
DFQ quantizes models without accessing real data~\cite{guo2022squant,choi2021qimera,li2023stableq,bai2023robustness,chen2024texq,qian2023adaptive,li2023hard,jeon2023genie,choi2022s,li2024genq}.
Most previous DFQ methods focus on CNN, where the BNS can be adopted as the regularization term~\cite{cai2020zeroq,zhang2021diversifying}. 
% For example, ZeroQ~\cite{cai2020zeroq} adopts BNS to regulate the distribution of synthetic data to be similar to the real data.
%
However, BNS is infeasible for ViTs built on the LN.
Recently, few efforts have been explored to accommodate ViTs~\cite{li2022patch,li2023psaqv2,ramachandran2024clamp,choi2024mimiq,husparse}.
PSAQ-ViT~\cite{li2022patch} introduces the first DFQ method for ViTs. They discover that Gaussian noise yields homogeneous patches, while the real image yields heterogeneous patches.
% Gaussian noise results in distinct multi-head self-attention (MHSA) module outputs compared with real images. Specifically, Gaussian noise yields homogeneous MHSA outputs, while the real image yields heterogeneous MHSA outputs.
%
Thus, patch similarity entropy (PSE) loss is proposed to optimize the Gaussian noise towards real-like images by making them showcase heterogeneous patches.
Based on PSAQ-ViT, PSAQ-ViT V2~\cite{li2023psaqv2} further introduces an adversarial learning strategy~\cite{GAN}. \cite{ramachandran2024clamp} incorporates contrastive learning and proposes an iterative generation-quantization PTQ-based DFQ method. 
\cite{husparse} proposes a sparse generation method to remove noisy and hallucination backgrounds in synthetic images. 
SynQ~\cite{KimKK25} introduces an overall framework for DFQ by integrating a low-pass filter, class activation map alignment, and difficult samples learning. 
For a more comprehensive DFQ survey, we invite the reader to see \cite{KimCLCK25}.

%
% Although some progress has been made, the performance of the current DFQ methods of ViTs is still less than expected.

\section{Method}

\subsection{Preliminaries}

\subsubsection{Quantizers}

We employ the linear quantizer for all weights and activations, except for the attention scores, which use a log2 quantizer to handle highly non-negative and uneven values~\cite{li2023repq,lin2022fqvit,frumkin2023jumping}. 
For the the linear quantizer, given a full-precision input $\mathbf{x}$ and bit-width $b$, the quantized value $\mathbf{x}_q$ and the de-quantized value $\Bar{\mathbf{x}}$ are computed as follows:
\begin{align}
\label{eq:linear_quant}
  \mathbf{x}_q  &= \text{clip}\left(\left\lfloor \frac{\mathbf{x}}{\Delta} \right\rceil + z, 0, 2^b-1 \right),  \Bar{\mathbf{x}} = \Delta \cdot \left(\mathbf{x}_q - z\right),
\end{align}
where $\left\lfloor\cdot\right\rceil$ denotes rounding to the nearest integer, and $\text{clip}$ limits the value to $[0, 2^b-1]$. Here, $\Delta$ and $z$ are the scale factor and zero-point, respectively.
For the log2 quantizer:
\begin{align}
\label{eq:log2_quant}
 \mathbf{x}_q &= \text{clip}\left(\left\lfloor -\log_2 \frac{\mathbf{x}}{\Delta} \right\rceil, 0, 2^b-1 \right),  \Bar{\mathbf{x}} = \Delta \cdot 2^{-\mathbf{x}_q}.
\end{align}

% We employ the popular linear quantizer for all weights and activations except for the attention scores. 
% %
% Specifically, given full-precision input $\mathbf{X}$ and bit-width $b$, 
% %
% the computation process of the quantized integer $\mathbf{X}_q$ and de-quantization value $\Bar{\mathbf{X}}$ in linear quantizer is defined as:
% %
% \begin{align}
% \label{eq:UQ}
%   \mathbf{X}_q  = \text{clip}\left(\left\lfloor \frac{\mathbf{X}}{\Delta} \right\rceil+z, 0, 2^b-1 \right), \, \Bar{\mathbf{X}} = \Delta\left(\mathbf{X}_q-z\right).
%   % \approx \mathbf{X},
% \end{align}
% %
% where $\left\lfloor\cdot\right\rceil$ denotes the round function that rounds the input to the nearest integer, $\text{clip}$ function clips the input between $0$ and $2^b-1$, $s$ and $z$ are the scale factor and the zero-point, respectively. 
% %
% Following~\cite{li2023repq,lin2022fqvit,frumkin2023jumping}, we employ the log2 quantizer to handle highly non-negative and uneven attention scores. The computation process of the quantized integer $\mathbf{X}_q$ and de-quantization $\Bar{\mathbf{X}}$ in log2 quantizer is defined as:
% % %
% \begin{align}
%  \mathbf{X}_q = \text{clip}\left(\left\lfloor -\log_2 \frac{\mathbf{X}}{\Delta} \right\rceil, 0, 2^b-1 \right), \, \Bar{\mathbf{X}} = \Delta \cdot 2^{-\mathbf{X}_q}.
%  %\approx \mathbf{X}.
% \end{align}

\subsubsection{Data Synthesis}

DFQ methods parameterize synthetic images and optimize them toward real-like images with a pre-trained full-precision model $F$. Given a image $\tilde{\bm{I}}$ initialized from Gaussian noise, the one-hot loss~\cite{GDFQ} is introduced to learm label-related semantics:
\begin{equation}
	\mathcal{L}^{\text{OH}}(\tilde{\bm{I}}) = CE(F(\tilde{\bm{I}}),c)
	\label{eq:l_oh}
\end{equation}
where $CE(\cdot, \cdot)$ represents the cross entropy, $c$ is a random class label, and $F(\cdot)$ returns the predicted probability for image $\tilde{\bm{I}}$. 

Moreover, total variance (TV)~\cite{yin2020dreaming} loss is a smoothing regularization term to improve the image quality:
\begin{equation}
	\mathcal{L}^{\text{TV}}(\tilde{\bm{I}}) = \iint |\nabla \tilde{\bm{I}}(\tau_1,\tau_2)|d\tau_1d\tau_2.
\label{eq:l_tv}
\end{equation}
where $\nabla \tilde{\bm{I}}(\tau_1,\tau_2)$ denotes the gradient at $\tilde{\bm{I}}$ at $(\tau_1,\tau_2)$.

To perform DFQ for ViTs, PSAQ-ViT~\cite{li2022patch} proposes patch similarity entropy (PSE) loss. It first compute patch similarity $\bm{\Gamma}_l[i,j] = \frac{u_i \cdot u_j}{||u_i|| ||u_j||}$, where $u_i, u_j$ are feature vectors of MHSA outputs in $l$-th block and $||\cdot||$ denotes the $l_2$ norm. 
% % %
Then, it estimates the density function $\hat{f_l}(x) = \frac{1}{Mh} \sum_{m=1}^{M} K\left(\frac{x-x_m}{h}\right)$, where $K(\cdot)$ is a normal kernel, $h$ is the bandwidth, and $x_m$ is the kernel center derived from $\bm{\Gamma}_l$. Finally, the PSE loss is defined as:
\begin{equation}
\mathcal{L}^{\text{PSE}}(\tilde{\bm{I}}) = \sum_{l=1}^{L} \int \hat{f_l}(x) \log \left[ \hat{f_l}(x) \right] dx,
\label{eq:pse}
\end{equation}
where $L$ is the block number of the model.

\subsection{Observations} 
\label{sec:Observations}

\begin{table}
\small
% \begin{table}[ht]
\centering
% \resizebox{\linewidth}{!}{
\begin{tabular}{c|c|c|c}
\toprule[1.25pt]
\diagbox{Classes}{Method} & Real &  PSAQ-ViT~\cite{li2022patch} & APA (Ours) \\
\midrule[0.75pt]
Class 1 &  0.68 &  0.44 & 0.64 \\ 
Class 2 &  0.32 &  0.26 & 0.31 \\ 
Class 3 &  0.41 &  0.31 & 0.36 \\
\bottomrule[1.0pt]
\end{tabular}
% }
\caption{
Average cosine similarity of three randomly selected classes. For real images, the similarity is measured within the class itself, while for PSAQ-ViT and APA, the similarity is measured between synthetic and real images of the same class. The results show that APA achieves higher similarity than PSE loss, indicating aligned semantics.
}
\label{tab:insight}
% \end{table}
\end{table}

%
% Both visualization and quantitative results reveal a pronounced distorted semantic, highlighting the issue of \textbf{semantic distortion}.

Existing DFQ methods have made significant progress. However, through carefully analyzing the synthetic images, we reveal that these images suffer from issues of semantic distortion and semantic inadequacy, both of which hinder further advancement in the DFQ of ViTs.

The \textbf{semantic distortion} issue refers to the significant divergence between the semantics of synthetic images and real images.
To demonstrate this, we visualize the features of synthetic and real images in Fig.\,\ref{fig:insight-tsne}. Note that the penultimate feature typically is regarded to represent the semantics of the input~\cite{zeiler2014visualizing,bengio2013representation,naseer2021intriguing,yosinski2015understanding}, 
It is clear that the features of PSAQ-ViT diverge significantly from real images, suggesting that these images fail to capture the true semantic distribution of real data.
Tab.\,\ref{tab:insight} quantitatively measures the semantics using the average cosine similarity (ranging from $-1$ to $1$). We also report the intra-class similarity within real images as an approximate upper bound for comparison.
This result further supports the observation of low similarity between synthetic images from PSAQ-ViT and real images. For instance, for Class 1, the intra-class similarity within real images is 0.68, while PSAQ-ViT achieves only 0.44.
% Similarly, for Class 3, while the real images exhibit a similarity of 0.41, PSAQ-ViT achieves only 0.31.
%
%
%

The \textbf{semantic inadequacy} issue refers to the presence of dull regions in synthetic images, which contain redundant or non-semantic content~\cite{husparse,kim2022dataset}, hindering the model's learning process. As indicated in \cite{choi2021qimera,chen2024texq}, a diverse content and textures generally suggests rich information. As shown in Fig.\,\ref{fig:insight-inadequacy}, many regions of synthetic images generated by PSAQ-ViT and PSAQ-ViT V2 exhibit a lack of diversity in content, with overly simplified textures. Specifically, the central region of PSAQ-ViT images only contains faint object structures, while PSAQ-ViT V2 images appear excessively smoothed and indistinct.

For high-bit quantization, where model capacity is largely retained~\cite{li2021brecq}, the performance degradation remains relatively minor even using semantic distorted and inadequate images~\cite{li2022patch,ramachandran2024clamp}.
However, in low-bit quantization, where model capacity is severely damaged and informative images are essential for recovering performance~\cite{zhong2022intraq,chen2024texq}, fine-tuning on these poor-quality images leads to inferior generalization to real datasets, resulting in limited performance.
For example, as shown in Tab.\,\ref{tab:ImageNet}, the W4A4 ViT-B fine-tuned on real images yields 68.16\%, whereas PSAQ-ViT only achieves 36.32\%.

\begin{figure}[!th]
\centering
\includegraphics[width=0.8\linewidth]{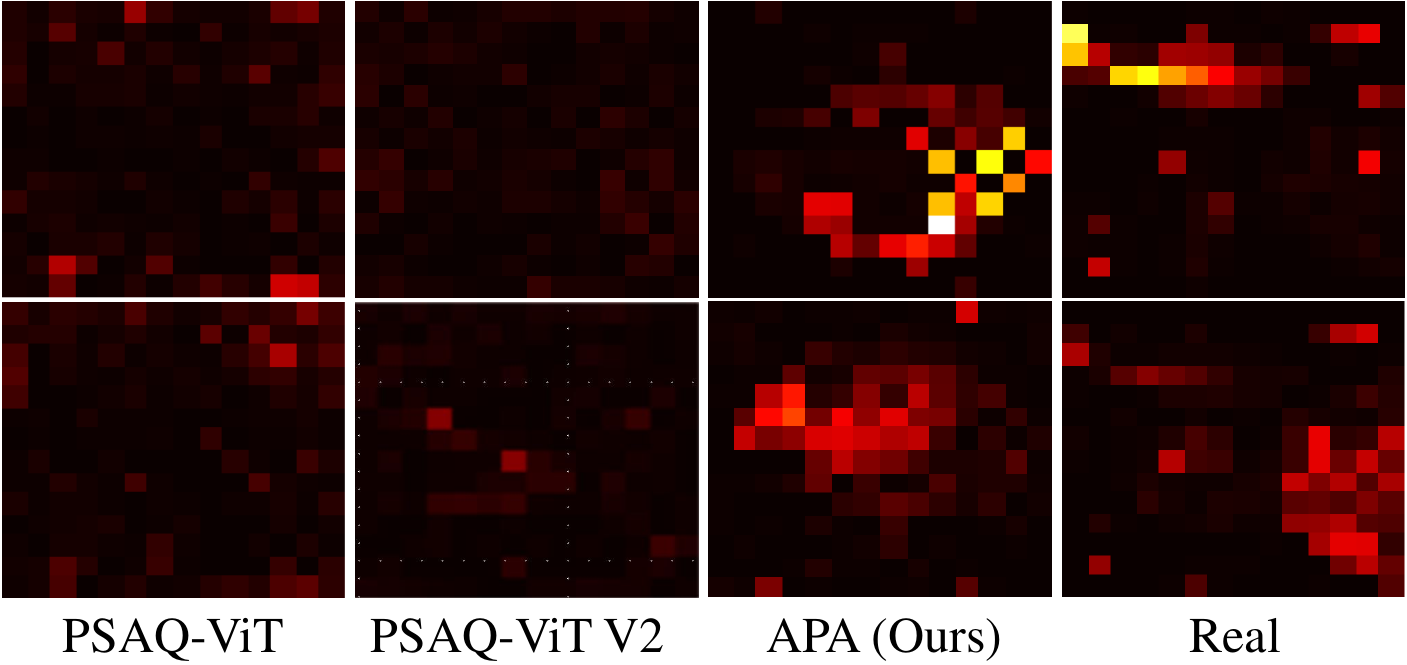}
\caption{Comparison between attention maps.}
\label{fig:attn-vis}
\vspace{-3mm}
\end{figure}

\subsection{Semantics Alignment and Reinforcement Data-Free Quantization}

In the following, we introduce the proposed SARDFQ, whose framework is illustrated in Fig.\,\ref{fig:framework}.

\subsubsection{Attention Priors Alignment}
\label{sec:APA}

% APA uses Gaussian Mixture Model (GMM) to generate diverse attention priors, simulating the semantic distribution characteristics of multiple objects/regions in real data (such as center periphery structure, multiple saliency peaks). In the optimization process, by minimizing the difference between the synthesized image attention map and the GMM prior (Equations 6-7), the generator is forced to improve in the following two aspects: 1) semantic focus: enhancing local region responses related to the target category (such as animal heads and vehicle contours), and suppressing irrelevant background noise; 2) Distribution diversity: By randomly sampling GMM components, covering different semantic combination patterns (such as multi object coexistence scenes). This constraint essentially distills ViT's hierarchical attention knowledge into the generation process, aligning the synthesized image with the semantic distribution of the real data in the feature space (as shown in the t-SNE visualization in Figure 1). 

%
% Specifically, the self-attention mechanism extracts and integrates semantic information by allocating more attention to tokens rich in semantics and less to tokens with minimal semantics~\cite{dosovitskiyimage,haurum2023tokens,chen2023cf}. 
%

In ViTs, the self-attention mechanism encodes semantic correlations between image regions, where high-response areas in attention maps strongly correlate with semantic-discriminative content~\cite{dosovitskiyimage,haurum2023tokens,chen2023cf}.
However, existing DFQ methods overlook this intrinsic property in the generation process.
As a result, as shown in Fig.\,\ref{fig:attn-vis}, synthetic images often exhibit disordered and unnatural attention patterns, with attention maps either overly diffuse or misaligned toward peripheral regions. This undermines their ability to preserve semantic-discriminative content, causing semantic distortion, as demonstrated in Fig.\,\ref{fig:insight-tsne} and Tab.\,\ref{tab:insight}.
%
% As a result, the generated images often exhibit disordered attention patterns (Fig. 3), where attention weights are either overly diffused or misallocated to non-informative regions.
%
In response, we propose Attention Priors Alignment (APA), which improves semantics alignment by optimizing synthetic images to follow randomly generated structure attention priors.

% APA模块通过显式约束合成图像的注意力分布与真实语义模式对齐，从根本上解决了现有方法中语义不足的问题。在ViT中，自注意力机制通过动态权重编码不同图像区域间的语义关联强度，高层注意力图直接反映了模型对关键语义特征的关注程度。APA采用高斯混合模型（GMM）生成多样化的注意力先验，模拟真实数据中多物体/多区域的语义分布特性（如中心-周边结构、多显著性峰值）。在优化过程中，通过最小化合成图像注意力图与GMM先验的差异（式6-7），迫使生成器在以下两方面改进：1）语义聚焦性：增强与目标类别相关的局部区域响应（如动物头部、车辆轮廓），抑制无关背景噪声；2）分布多样性：通过随机采样GMM成分，覆盖不同语义组合模式（如多物体共存场景）。这种约束本质上是将ViT的层级注意力知识蒸馏到生成过程中，使得合成图像在特征空间中与真实数据的语义分布对齐（如图1的t-SNE可视化所示）。实验表明，APA使合成图像与真实数据的注意力相似度提升41%（表1），且高层特征余弦相似度增加29%，验证了其对语义增强的核心作用。

Specifically, given a synthetic image $\tilde{\bm{I}}$, we first obtain its attention maps in the $h$-th head of the $l$-th block, denoting as $\mathbf{A}_{l, h} \in \mathbb{R}^{N \times N}$, where $N$ represents the total number of tokens. 
In DeiT, the attention of the classification token toward other tokens serves as the indicator for semantic versus non-semantic parts~\cite{bolya2022token}. Thus, we extract $\mathbf{A}^c_{l, h} \in \mathbb{R}^{1 \times (N-1)}$ from $\mathbf{A}_{l, h}$, representing the attention of the classification token to all tokens except itself. 
We then randomly generate attention priors $\tilde{\mathbf{A}}_{l, h}$, whose generation is detailed in the next part, and align $\mathbf{A}^c_{l, h}$ with $\tilde{\mathbf{A}}_{l, h}$ by:
\begin{align}
    \mathcal{L}_{l, h}(\tilde{\bm{I}}) = \text{MSE}(\mathbf{A}^c_{l, h} - \Tilde{\mathbf{A}}_{l, h}),
\label{eq:apa1}
\end{align}
where $\text{MSE}$ represents the mean squared error.
For Swin models that do not use a classification token, we substitute $\mathbf{A}^c_{l, h}$ in Eq.\,\ref{eq:apa1} with the average attention map of all tokens~\cite{chen2023cf}.
As noted in~\cite{cordonnier2020relationship}, ViTs initially focus on all regions to capture low-level information in shallow blocks and gradually shift their focus toward semantic regions in deeper blocks to extract high-level semantic information. Leveraging this property, we selectively apply $\mathcal{L}^{\text{APA}}_{l, h}$ to deeper blocks, progressively aligning attention towards semantically relevant areas. The total APA loss is computed as a depth-weighted sum of the individual Eq.\,\ref{eq:apa1} across these deeper blocks:
\begin{align}
   \mathcal{L}^{\text{APA}}(\tilde{\bm{I}}) =  \sum_{l=S}^L  \sum_{h=1}^H \frac{l}{L} \mathcal{L}_{l, h}(\tilde{\bm{I}}),
\label{eq:apa}
\end{align}
where $S$ is a pre-given hyper-parameter denoting the start of deep blocks and are experimentally set $S = \frac{L}{2}$.

% \footnote{xxx In the appendix, we provide the comparison of attention maps to further validate the effect of APA loss.}.

\begin{figure}[!t]
\centering
\includegraphics[width=0.8\linewidth]{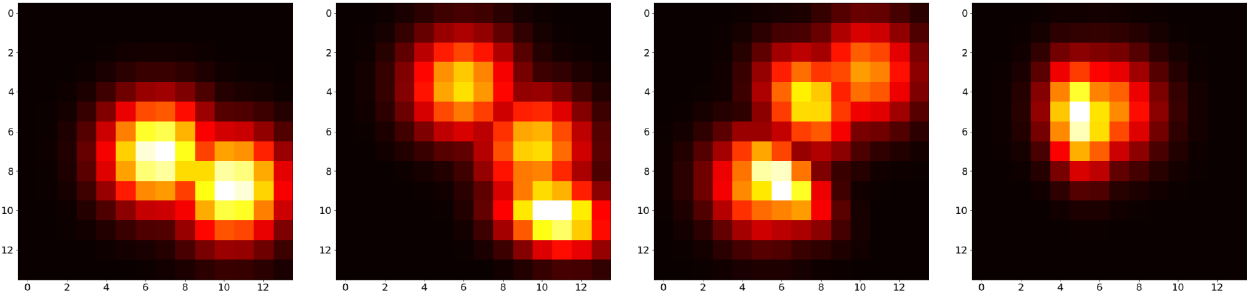}
\caption{Examples of generated attention priors.}
\label{fig:attn_p}
\vspace{-3mm}
\end{figure}

\textbf{Attention Priors Generation}. 
To generate attention priors $\tilde{\mathbf{A}}_{l, h}$, Gaussian Mixture Models (GMMs) are employed as it is the most commonly used distribution with high flexibility. 
%
%
% \begin{wrapfigure}{r}{4cm} % 靠文字内容的左侧
% \centering
% \includegraphics[width=\linewidth]{attn_p.pdf}\
% \caption{Examples of generated attention priors.}
% \label{fig:attn_p}
% \end{wrapfigure}
%
%
ViTs utilize different attention heads to capture diverse patterns and learn varied, informative representations~\cite{chefer2021generic}. Thus, we use distinct GMMs for each head. Note that the goal here is not to replicate real attention maps precisely, but to generate simulated structure attention priors to guide the learning of synthetic images.

In particular, we first initialize an all zero matrix $\Tilde{\mathbf{P}} \in \mathbb{R}^{H \times W}$, 
where $H=W=\sqrt{N-1}$ for DeiT, $H=W=\sqrt{N}$ for Swin. For example, for DeiT-S, the $H=W=\sqrt{196}=14$. Then, we generate $k$ two-dimensional Gaussian distributions, where $k$ is randomly sampled from $1 \sim K_{APA}$ and $K_{APA}$ is set to $5$ in all experiments. Each Gaussian has its mean and covariance\footnote{The pseudo code is detailed in appendix.}. Consequently, the matrix element at the $i$-th row and $j$-th column, $\tilde{\mathbf{P}}[i,j]$, is determined by:
\begin{align}
    \Tilde{\mathbf{P}}[i,j] = \max_{m=1, \cdots, k} \mathbf{G}^m[i,j].
\end{align} 
Then, $\Tilde{\mathbf{P}}$ is normalized by:
\begin{align}
    \Tilde{\mathbf{P}}_n = \frac{\Tilde{\mathbf{P}}}{\sum \Tilde{\mathbf{P}}} \cdot (1 - x).
\end{align} 

Here, for DeiT that incorporates the classification token, $x$ is randomly sampled from a uniform distribution $U(0, 1)$, representing the proportion of the attention score that the classification token allocates to itself. For Swin, which does not use a classification token, $x$ is set to 0.
Finally, $\Tilde{\mathbf{P}}_n$ is flatten to match the dimensionality:
\begin{align}
   \Tilde{\mathbf{A}}_{l, h} = \text{flatten}(\Tilde{\mathbf{P}}_n).
\end{align} 

Fig.\,\ref{fig:attn_p} displays examples of generated attention priors.

\textbf{Discussion.} 
APA prevents attention disorder, ensuring that synthetic images exhibit more coherent and natural attention patterns, as demonstrated in Fig.\,\ref{fig:attn-vis}. As a result, APA selectively enhances the responses in certain regions, effectively emphasizing semantic-discriminative regions within synthetic images and thus prompting the discriminative features. Although simple, APA enables synthetic images to align better with the real semantics, as validated by both visual and quantitative evaluations. As shown in Fig.\,\ref{fig:insight-tsne}, compared to PSAQ-ViT, features obtained after applying the APA loss are more closely aligned with real images, indicating better semantic alignment. The quantitative results in Tab.\,\ref{tab:insight} further support that APA achieves superior semantic alignment. For example, in Class 1, the intra-class similarity for PSAQ-ViT is 0.44, whereas APA achieves a higher similarity of 0.64.

\subsubsection{Multi-Semantic Reinforcement}
\label{sec:MSR}

Current DFQ methods for ViTs~\cite{li2022patch,li2023psaqv2,ramachandran2024clamp} optimize images through global optimization, treating the entire image as a single semantic unit.
However, this is affected by low-rank structural regularity~\cite{huang1999statistics}, where adjacent pixels exhibit strong similarity, leading to dull regions with redundant or non-semantic content for model learning~\cite{kim2022dataset}.
This issue is further exacerbated by the tokenization mechanism in ViTs, as processing images in fixed-size patches make dull regions increase at the patch level~\cite{dong2021attention,wang2022anti}.
Consequently, as shown in Fig.\,\ref{fig:insight-inadequacy}, synthetic images generated by existing methods exhibit large dull regions, resulting in semantic inadequacy~\cite{kim2022dataset}.
In response, we propose Multi-Semantic Reinforcement (MSR), which applies localized patch optimization to enhance semantic richness by learning local patches with distinct semantics.

Specifically, for a synthetic image $\tilde{\bm{I}}$, instead of feeding only the entire image, we also feed its patches and optimize them individually. Initially, we select $m$ non-overlapping patches, where $m$ is chosen randomly from the set $\{1, 2, \ldots, K_{MSR}\}$, with $K_{MSR}$ set to 4 in all experiments. These $m$ patches are then cropped and resized to match the model’s input dimensions:
\begin{equation}
    \{ \tilde{\bm{I}}^{i}_{MSR} \}_{i=1,\cdots,m} =  \text{resize}\big(\text{crop}_{m}(\tilde{\bm{I}})\big),
\end{equation}
where $\text{crop}_{m}(\cdot)$ crops $m$ non-overlapping patches from input, and $\text{resize}(\cdot)$ is the resize function. Each patch, denoted as $\tilde{\bm{I}}^{i}_{MSR}$, is treated as a new image with an assigned semantic target $c^{i}$.
% encouraging varies semantics in $\tilde{\bm{I}}$.
%
% For simplicity, the loss function used to learn label-related semantics in both $\tilde{\bm{I}}^{i}_{MSR}$ and $\tilde{\bm{I}}$ is defined in Sec.\,\ref{sec:Softlabel}.
%
Note that the gradient is backpropagated only to update the corresponding patch in the original image, leaving the rest of the image unaffected.

\textbf{Softlabel Learning.} 
%
%
% optimizes synthetic images until their confidence for the target class approaches 1. 
%
% However, this leads to overfitting on a specific class, limiting semantic diversity~\cite{zhong2022intraq}. Moreover, Eq.\,\ref{eq:l_oh} 
The one-hot loss (Eq.\,\ref{eq:l_oh}) only learns the semantics of the target class, making it unsuitable for $\tilde{\bm{I}}$ under MSR, as its patches $\{ \tilde{\bm{I}}^{i}_{MSR} \}_{i=1,\cdots,m}$ contain distinct semantics.
In response, we propose Softlabel Learning (SL), which applies multiple semantic targets to accommodate the learning of images augmented by MSR.
Specifically, we first sample $Z \in \mathbb{R}^C \sim U(0, 1)$, then modify its values by:
 \[
 \begin{cases}
 Z[c^{i}] \sim U(\epsilon_1, \epsilon_2), & \text{for } \tilde{\bm{I}}^{i}_{MSR}, \\
 Z[c^{1}, \ldots, c^{m}] \sim U(\epsilon_1, \epsilon_2), & \text{for } \tilde{\bm{I}},
 \end{cases}
 \]
where $U(\epsilon_1, \epsilon_2)$ denotes the uniform distribution over the interval $[\epsilon_1, \epsilon_2]$, $m$ is the number of patches determined in MSR, and $\epsilon_1$ and $\epsilon_2$ control the softness, both empirically set consistently to 5 and 10 in all experiments.
The soft target is defined as $T_s = \text{softmax}(Z)$, and SL loss is:
\begin{equation}
{\cal L}^{\text{SL}}(\tilde{\bm{I}}/\tilde{\bm{I}}^{i}_{MSR}) = SCE\big(F(\tilde{\bm{I}}/\tilde{\bm{I}}^{i}_{MSR}), T_s \big),
\label{eq:l_SL}
\end{equation}
where $SCE(\cdot, \cdot)$ is soft cross entropy and $F(\cdot)$ returns the predicted probability for its input. SL facilitates smooth learning between across semantic targets, ensuring that MSR-enhanced images receive consistent supervision rather than conflicting supervision.

\textbf{Discusion.} 
%
% By leveraging localized patch optimization, MSR effectively reduces dull regions and leans rich semantics in synthetic images, as validated by the greater and diverse content and textures in Fig.\,\ref{fig:insight-inadequacy}. On the one hand, cropped patches of $\tilde{\bm{I}}$ are leaned to contain semantics, reducing the space for dull regions.
%
By leveraging localized patch optimization, MSR ensures that each patch contributes unique semantics, forcing synthetic images to capture diverse features rather than being dominated by large homogeneous dull regions.
As demonstrated by the richer and more diverse content and textures in Fig.\,\ref{fig:insight-inadequacy}, MSR effectively reduces dull regions and enhances semantic richness in synthetic images.
Moreover, MSR transforms synthetic images into composites of multiple semantic objects rather than a single unit, thereby providing more distinct semantic samples, \emph{i.e.}, $\{ \tilde{\bm{I}}^{i}_{MSR} \}_{i=1,\cdots,m}$, for model training.
Unlike traditional cropping used in data augmentation, which aims to improve classification robustness by training with cropped patches labeled with the original class, MSR aims for semantic richness within synthetic images, ultimately enabling accurate data-free quantization.

\subsection{Overall Pipeline}

The overall pipeline consists of two stages: data synthesis and quantized network learning. The first stage uses the proposed SARDFQ to produce synthetic images. The second stage fine-tunes the quantized model using the generated synthetic images.

\subsubsection{Data Synthesis}

In the data synthesis stage, we combine the proposed APA loss of Eq.\,\ref{eq:apa}, SL loss of Eq.\,\ref{eq:l_SL}, and TV loss of Eq.\,\ref{eq:l_tv} to formulate the objective function as follows:
\begin{equation}
	\begin{aligned}
   \mathcal{L}_{G}(\tilde{\bm{I}}) & = \alpha_1 \mathcal{L}^{\text{APA}}(\tilde{\bm{I}}) + {\cal L}^{\text{SL}}(\tilde{\bm{I}}) + 0.05 \mathcal{L}^{\text{TV}}(\tilde{\bm{I}}).
	\end{aligned}
\label{eq:l_g}
\end{equation}
where $\alpha_1$ is hyperparameters and is determined by grid-search. Note that the weight of TV loss is fixed to 0.05, following \cite{li2022patch}, to avoid a cumbersome hyperparameter search.

\subsubsection{Quantized Network Learning}

Recently DFQ methods have introduced the PTQ methods in learning quantized models due to their advantages of speed, memory efficiency, and performance~\cite{jeon2023genie,ramachandran2024clamp}. Thus, following the success of \cite{li2021brecq,wei2021qdrop}, we fine-tune the quantized network block-wisely. Specifically, denote $\mathbf{X}_l$ as the outputs of the $l$-th block of the full-precision model, and $\Bar{\mathbf{X}}_l$ represent outputs of the quantized counterpart. The reconstruction loss is defined as:
\begin{align}
  \mathcal{L}_l = \| \mathbf{X}_l - \Bar{\mathbf{X}}_l \|_2.
\end{align}

Here, $\mathcal{L}_l$ is only backward to update weights within $l$-th block. Note that for a fair comparison, all compared methods adopt the same quantized network learning stage.

\section{Experiment}
\label{sec:experiment}

\subsection{Implementation Details}

\textbf{Models and Tasks.} 
We evaluate the performance of SARDFQ by test quantized ViT-S/B~\cite{dosovitskiyimage}, DeiT-T/S/B~\cite{touvron2021training}, and Swin-S/B~\cite{liu2021swin} on the classification task using ImageNet~\cite{russakovsky2015imagenet}. The pre-trained models are downloaded from the timm library. In appendix, we further provide results on detection and segmentation tasks.
%

% \footnote[1]{https://github.com/rwightman/pytorch-image-models.}

\begin{table*}[ht]
\centering
\begin{tabular}{c|c|c|c|c|c|c|c}
    \toprule[1.25pt]
    Model & W/A & Real & Gaussian noise & PSAQ-ViT~\cite{li2022patch} & PSAQ-ViT V2~\cite{li2023psaqv2} & SMI~\cite{husparse}  & SARDFQ (Ours) \\
    \midrule[0.75pt]

    \multirow{3}*{\tabincell{c}{ViT-S \\ (81.39)}} & 4/4 & 66.57 & 6.02 & 47.24 & 41.53 & 24.33$_{29.41}$& \textbf{50.32} \\
                                 & 5/5 & 76.69 & 36.77 & 71.59 & 68.41 & 61.33$_{65.19}$ & \textbf{74.31} \\
                                 & 6/6 & 79.46 & 61.20 & 77.20 & 74.76 & 72.95$_{72.46}$ &  \textbf{78.40} \\

    \cmidrule{1-8}

    \multirow{3}*{\tabincell{c}{ViT-B\\ (84.54)}} & 4/4 & 68.16 & 0.15 & 36.32  & 26.32 & 35.27$_{19.67}$ & \textbf{51.84} \\
                                 & 5/5 & 79.21 & 4.16 & 68.48 & 67.95 & 67.53$_{57.13}$ & \textbf{70.70} \\
                                 & 6/6 & 81.89 &  55.18 & 76.65 & 71.87 & 76.33$_{69.82}$ & \textbf{79.16} \\

    \cmidrule{1-8}

    \multirow{3}*{\tabincell{c}{DeiT-T \\ (72.21)}} & 4/4 & 56.60 & 17.43 & 47.75 & 30.20 & 30.14$_{13.18}$ & \textbf{52.06} \\
                                  & 5/5 &  67.09 & 43.49 & 64.10 & 55.16 &  56.44$_{39.35}$ & \textbf{66.41} \\
                                  & 6/6 & 69.81 & 56.23 & 68.37 & 62.77 & 64.03$_{44.39}$ & \textbf{69.73} \\

    \cmidrule{1-8}

    \multirow{3}*{\tabincell{c}{DeiT-S \\ (79.85)}} & 4/4 & 68.46 & 20.89 & 58.28 & 45.53 & 42.77$_{11.71}$ & \textbf{62.29} \\
                                  & 5/5 & 75.06 & 41.06 & 71.90 & 63.14 & 62.88$_{29.13}$  & \textbf{74.06} \\
                                  & 6/6 & 77.87 & 65.63 & 75.85 & 68.85 & 71.65$_{37.69}$ & \textbf{77.31} \\

    \cmidrule{1-8}

    \multirow{3}*{\tabincell{c}{DeiT-B \\(81.85)}} & 4/4 & 77.07 & 47.20 & 71.75  & 66.43 & 65.33$_{59.04}$ & \textbf{72.17} \\
                                  & 5/5 & 79.86 &  65.46 & 78.45 & 76.77 & 76.74$_{75.33}$ & \textbf{78.72} \\
                                  & 6/6 & 80.90 & 62.79 & 80.00 & 79.22 & 78.81$_{77.66}$ & \textbf{80.15} \\

    \cmidrule{1-8}

    \multirow{3}*{\tabincell{c}{Swin-S \\(83.20)}} & 4/4 & 78.12 & 31.92 & 73.19 & 65.55 & 65.85 & \textbf{74.74} \\
                                  & 5/5 & 80.51 & 52.10 &  78.15 & 74.37 & 75.41 & \textbf{79.56} \\
                                  & 6/6 & 80.60 & 65.66 & 79.74  & 78.50 & 78.25 & \textbf{80.56} \\

    \cmidrule{1-8}

    \multirow{3}*{\tabincell{c}{Swin-B \\ (85.27)}} & 4/4 & 78.80 & 30.14 &  71.84 & 67.42 & 65.23 & \textbf{76.42} \\
                                  & 5/5 & 82.51 & 35.28 & 78.50  & 77.20 & 75.25 & \textbf{80.82} \\
                                  & 6/6 & 82.64 & 67.37 & 82.00  & 81.41 &  80.30 & \textbf{83.03} \\

    \bottomrule[1.0pt]
\end{tabular}
\caption{
% Quantization results on the ImageNet dataset. ``W/A'' indicates the bit-width of weights and activations, and ``Top-1'' represents the Top-1 accuracy (\%) of the quantized models. ``Real'' refers to using real images, ``No Data'' indicates the absence of real data during quantized network training, and ``SARDFQ (Ours)'' is the proposed method.
%
Quantization results on ImageNet dataset, with top-1 accuracy (\%) reported. The performance of the full-precision model is listed below the model name. ``W/A" denotes the bit-width of weights/activations. ``Real'' refers to using real images. For SMI~\cite{husparse}, we provide the performance of using dense (normal-sized numbers) and sparse (smaller-sized numbers) synthetic images, respectively. Note that for Swin models, we do not provide the results for sparse synthetic images as the sparse generation method of SMI is infeasible. 
}
\label{tab:ImageNet}
\vspace{-3mm}
\end{table*}

\textbf{Comparison methods.}
We compare our SARDFQ against Gaussian noise, real images, and previous methods including SMI~\cite{husparse} and PSAQ-ViT~\cite{li2022patch} and its subsequent version, PSAQ-ViT V2~\cite{li2023psaqv2}. 
For a fair comparison, we generate synthetic images using their methods and apply our quantized network learning strategy. We use the official code for SMI and PSAQ-ViT to reproduce their images, while PSAQ-ViT V2 is re-implemented by us, as no official code is available.\footnote{In appendix, we present practical efficiency results and more performance comparisons with more methods such as CLAMP-ViT~\cite{ramachandran2024clamp} on W8/A8 and W4/A8 settings.}

% Note that PSAQ-ViT and PSAQ-ViT V2 adopt different quantized network learning strategies. PSAQ-ViT only performs quantization parameter calibration while PSAQ-ViT V2 optimizes the network. 

\textbf{Experimental settings.}
All experiments were conducted using the PyTorch framework~\cite{paszke2019pytorch} on a single NVIDIA 3090 GPU.
In the data synthesis stage, synthetic images were initialized with standard Gaussian noise, generating 32 images in total. The Adam optimizer~\cite{kingma2014adam} with $\beta_1=0.5$, $\beta_2=0.9$ was used, with learning rates of 0.25 for Swin and 0.2 for others, and a total of 1,000 iterations.
For all models, $K_{APA}$, $K_{MSR}$, $\epsilon_1$, and $\epsilon_2$ were set to 5, 4, 5, and 10, respectively, based on a search with W4A4 DeiT-S. The value of $\alpha_1$ was determined by grid search for each model: 1e5 for DeiT-T/S, 1e4 for DeiT-B, 100 for ViT-B and Swin-B, 10 for Swin-S, and 1 for ViT-S.
Although further hyperparameter search may improve performance, the current settings already yield superior results.
In the quantized network learning stage, following \cite{zhong2023s}, the Adam optimizer with $\beta_1=0.9$, $\beta_2=0.999$ was used, with weight decay set to 0 and an initial learning rate of 4e-5, adjusted via cosine decay for 100 iterations. 
% Quantization parameters were determined by searching for percentile values that minimize quantization error~\cite{li2023repq}. We followed the two-stage optimization strategy from \cite{zhong2023s}.
%
A channel-wise quantizer was used for weights, and a layer-wise quantizer for activations, with all matrix multiplications in ViTs quantized~\cite{li2023psaqv2,li2022patch,li2023repq}.

\subsection{Quantization Results}

The quantization results are presented in Tab.\,\ref{tab:ImageNet}. Our SARDFQ demonstrates consistent improvements across various quantization bit-width configurations, particularly with low bit-width settings.
Specifically, for ViT-S, SARDFQ improves the performance by 3.08\% in the W4/A4 setting, 2.72\% in the W5/A5 setting, and 1.20\% in the W6/A6 setting. For ViT-B, SARDFQ achieves performance gains of 15.52\% in the W4/A4 setting, 2.22\% in the W5/A5 setting, and 2.51\% in the W6/A6 setting.
Results on DeiT also demonstrate the effectiveness of SARDFQ.
For example, on DeiT-T, SARDFQ shows a marked improvement by increasing top-1 accuracy by 4.31\% in the W4/A4 setting, 2.31\% in the W5/A5 setting, and 1.36\% in the W6/A6 setting. 
For DeiT-S, SARDFQ enhances top-1 accuracy by 4.01\% in the W4/A4 setting, 2.16\% in the W5/A5 setting, and 1.46\% in the W6/A6 setting. 
%
% As for DeiT-B, SARDFQ enhances top-1 accuracy by 0.42\% in the W4/A4 setting, 0.27\% in the W5/A5 setting, and 0.15\% in the W6/A6 setting.
%
The quantization results of Swin-S/B also affirm the superiority of our SARDFQ in enhancing model accuracy under different quantization configurations.
In particular, for Swin-S, the proposed SARDFQ increases the accuracy by 1.55\% for the W4/A4 setting, 1.41\% for the W5/A5 setting, and 0.82\% for the W6/A6 setting, respectively.
When it comes to Swin-B, the proposed SARDFQ increases the accuracy by 4.58\% for the W4/A4 setting, 2.32\% for the W5/A5 setting, and 1.03\% for the W6/A6 setting, respectively.

\subsection{Ablation Study}

All ablation studies are conducted on the W4A4 DeiT-S.

\begin{table}[h!]
\centering
\begin{tabular}{c|c|c|c}
    \toprule[1.25pt]
    \textbf{APA} & \textbf{MSR} & \textbf{SL} & \textbf{Acc. (\%)} \\
    \midrule[0.75pt]
    \multicolumn{3}{c|}{Baseline} &  51.73 \\
    \midrule[0.75pt]
    \checkmark & & & 60.26  \\
    & \checkmark & &   50.75 \\
    & & \checkmark &  52.02 \\
    \checkmark & \checkmark & & 61.58 \\
    \checkmark & & \checkmark &  60.51 \\
    & \checkmark & \checkmark & 56.08 \\
    \checkmark & \checkmark & \checkmark & \textbf{62.29} \\
    \bottomrule[1.0pt]
\end{tabular}
\caption{Influence of the proposed APA, MSR, and SL on accuracy. The baseline adopts the one-hot loss.}
\label{tab:components}
\vspace{-0.4em}
\end{table}

\textbf{Analysis of APA, MSR, and SL.}
%
% We first analyze the effectiveness of the proposed APA (Sec.\,\ref{sec:APA}), MSR (Sec.\,\ref{sec:MSR}), and SL (Eq.\,\ref{eq:l_SL}). Experimental results are present in Tab.\,\ref{tab:components}. When the APA and SL are individually added to synthesize images, the accuracy increases compared with the baseline. 
% %
% Notably, APA significantly boosts the baseline from 51.73\% to 60.26\%, supporting the analysis in Sec.\,\ref{sec:APA} that demonstrates the APA’s effectiveness in aligning semantics.
% %
% Applying MSR alone slightly reduces accuracy from 51.73\% to 50.75\%, indicating that one-hot loss is unsuitable for synthetic images augmented with MSR.
% %
% In contrast, when both MSR and SL are applied, performance rises to 56.08\%, suggesting that SL is more compatible with MSR than the one-hot loss.
% %
% Combining APA with either MSR or SL further increases performance. For example, applying both APA and SL presents an accuracy of 60.51\%. When all of the three strategies are applied, the best performance of 62.29\% can be obtained.
%
We analyze the effectiveness of the proposed APA (Sec.,\ref{sec:APA}), MSR (Sec.,\ref{sec:MSR}), and SL (Eq.,\ref{eq:l_SL}) in Tab.,\ref{tab:components}. Adding APA and SL individually to the baseline increases accuracy. Notably, APA boosts performance from 51.73\% to 60.26\%, confirming its effectiveness in aligning semantics (Sec.,\ref{sec:APA}).
Applying MSR alone slightly decreases accuracy from 51.73\% to 50.75\%, indicating that one-hot loss is unsuitable for MSR-augmented synthetic images.
However, when both MSR and SL are applied, accuracy rises to 56.08\%, suggesting SL is more compatible with MSR than one-hot loss.
Combining APA with either MSR or SL further improves performance. For example, APA and SL together yield an accuracy of 60.51\%, and when all three strategies are used, the best performance of 62.29\% is achieved.

\textbf{Analysis of priors distribution.}
Tab.\,\ref{tab:abl-type} showcases the results of using other distributions to formulate the attention priors. The unevenly distributed GMM and Laplace present comparable performance of 62.29\% and 62.16\%, respectively. 
% However, the uniform distribution hurts performance severely since attentions are evenly allocated. 
Moreover, GMM provides a similar performance to the real's, indicating it performs well in imitating the patterns of real images.

\begin{figure}[!th]
\centering
\begin{subfigure}{0.48\linewidth}
    \centering
    \includegraphics[width=\linewidth]{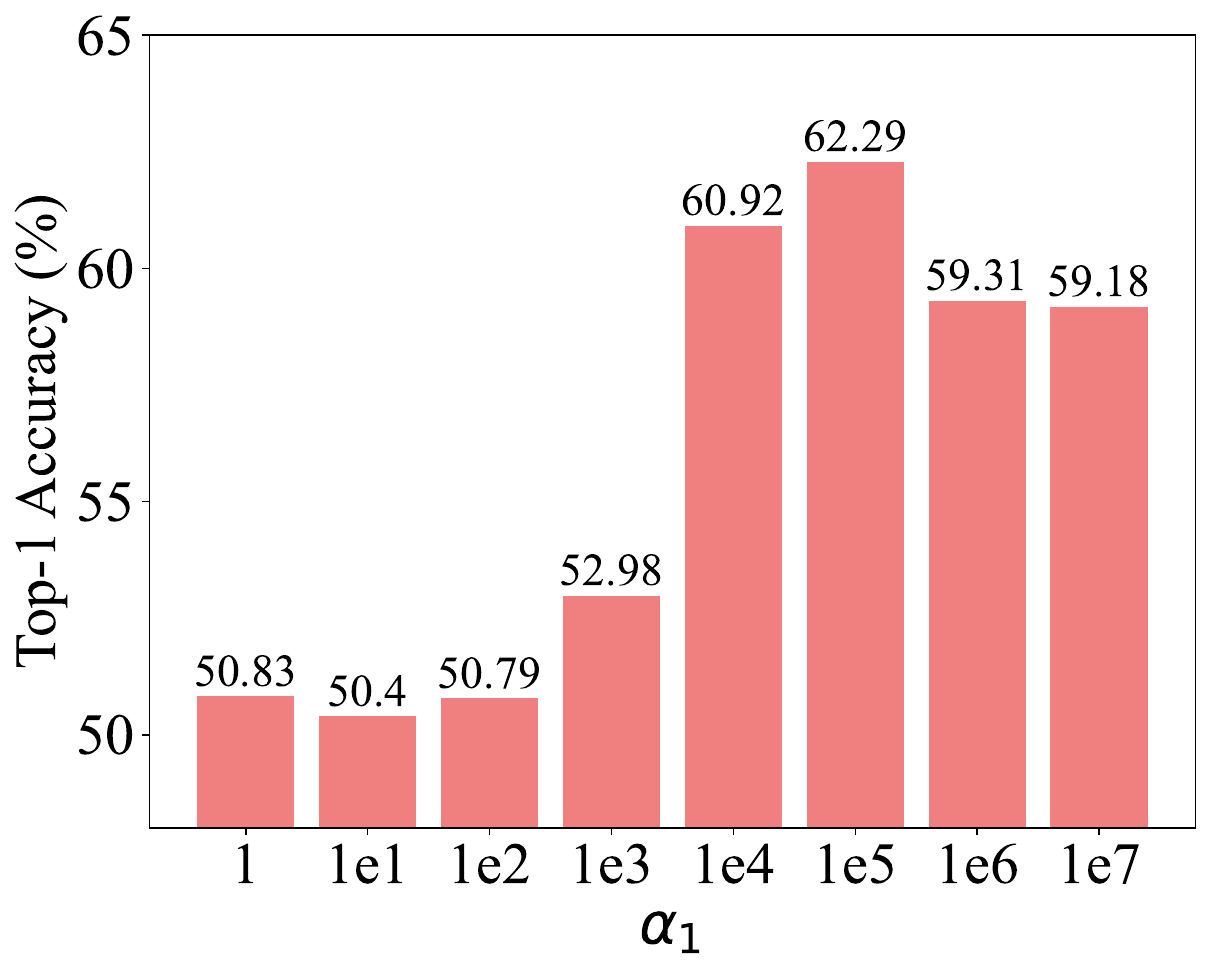} 
     % \caption{Ablation study of $\alpha_1$.}
     \caption{}
     \label{fig:coe-alpha1}
\end{subfigure}
\begin{subfigure}{0.5\linewidth}
\centering
    \includegraphics[width=\linewidth]{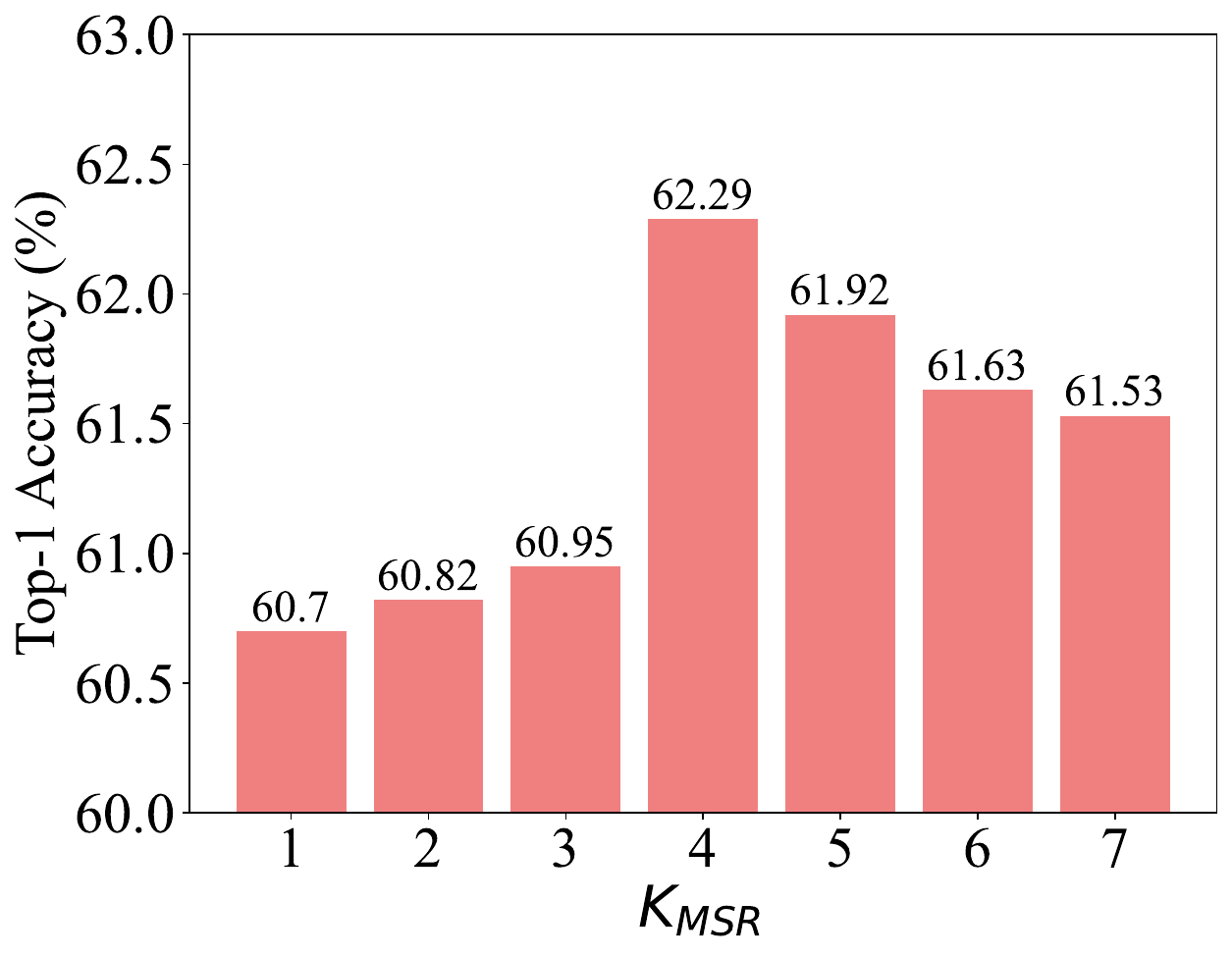} 
         % \caption{Ablation study of $K_{MSR}$.}
         \caption{}
         \label{fig:msr-KMSR}
\end{subfigure}
\caption{Effect of varying (a) $\alpha_1$ and (b) $K_{MSR}$.} 
\end{figure}

\textbf{Analysis of $\alpha_1$, $K_{APA}$, and $K_{MSR}$.%, and $\epsilon_1$, and $\epsilon_2$.
}
The $\alpha_1$ from Eq.\,\ref{eq:l_g} balance the importance of the proposed APA loss during the update of the synthetic images. 
Fig.\,\ref{fig:coe-alpha1} demonstrates that the optimal performance is achieved when $\alpha_1 = 1e5$. 
Incrementally increasing $\alpha_1$ improves performance up to 62.29\% at $\alpha_1 = 1e5$. However, further increases in $\alpha_1$ subsequently degrade performance.
The $K_{APA}$ in APA is the upper limit on the number of the Gaussian distributions used for priors generation. Tab.\,\ref{tab:abl-Kapa} displays the ablation study for different values of $K_{APA}$. The best accuracy is achieved when $K_{APA} = 5$.
The $K_{MSR}$ in MSR is the upper limit on the number of patches.
Fig.\,\ref{fig:msr-KMSR} demonstrates that when $K_{MSR}=4$, the optimal performance is achieved.
Using $K_{MSR}$ larger than 4 will hurt the accuracy. We consider this due to limited patch resolution if using a too large $K_{MSR}$.

% \begin{table}
% \begin{minipage}[t]{0.45\linewidth} % 调整宽度
% \centering
% \begin{tabular}{cc}
% \toprule[1.25pt]
% Priors Distribution  &  \textbf{Top-1} \\
% \midrule[0.75pt]
%  GMM & 62.29 \\
%  Laplace  & 62.16 \\
%    Real  & \textbf{63.19} \\
% \bottomrule[1.0pt]
% \end{tabular}
% \caption{Effect of priors type.}
% \label{tab:abl-type}
% \end{minipage}
% \hfill
% \begin{minipage}[t]{0.45\linewidth} % 设置宽度为文本宽度的一部分
% \centering
% \begin{tabular}{cc}
% \toprule[1.25pt]
% \textbf{$K_{APA}$}  &\textbf{Top-1} \\
% \midrule[0.75pt]
%  1  &  61.13  \\
%  3  &  61.52 \\
%  5  &  \textbf{62.29}  \\
%  7  &  61.53 \\
%  9  &  61.05 \\
% \bottomrule[1.0pt]
% \end{tabular}
% \caption{Effect of $K_{APA}$.}
% \label{tab:abl-Kapa}
% \end{minipage}
% %
% \end{table}

% \begin{table}
% %
% \begin{minipage}[t]{0.45\linewidth}
% \centering
% \begin{tabular}{cc}
% \toprule[1.25pt]
% S &  Top-1 \\
% \midrule[0.75pt]
% $0$ & 61.96  \\
% $L/2$ & \textbf{62.29} \\
% \bottomrule[1.0pt]
% \end{tabular}
% \caption{Effect of $S$.}
% \label{tab:abl-eq7-S}
% \end{minipage}
% \hfill
% %
% \begin{minipage}[t]{0.45\linewidth} % 调整宽度
% \centering
% \begin{tabular}{cc}
% \toprule[1.25pt]
% w. $\frac{l}{L}$ &  Top-1 \\
% \midrule[0.75pt]
%  \checkmark & \textbf{62.29} \\
%  $\times$  & 61.32 \\
% \bottomrule[1.0pt]
% \end{tabular}
% \caption{Effect of $\frac{l}{L}$.}
% \label{tab:abl-eq7-scale}
% \end{minipage}
% \vspace{-0.4em}
% \end{table}

\begin{table}[t]
\centering
% 第一行子表格
\begin{subtable}[t]{0.45\linewidth}
\centering
\begin{tabular}{cc}
\toprule[1.25pt]
Priors Distribution  &  \textbf{Top-1} \\
\midrule[0.75pt]
 GMM & 62.29 \\
 Laplace  & 62.16 \\
   Real  & \textbf{63.19} \\
\bottomrule[1.0pt]
\end{tabular}
% \subcaption{Effect of priors type.}
\subcaption{}
\label{tab:abl-type}
\end{subtable}
\hfill
\begin{subtable}[t]{0.45\linewidth}
\centering
\begin{tabular}{cc}
\toprule[1.25pt]
\textbf{$K_{APA}$}  &\textbf{Top-1} \\
\midrule[0.75pt]
 1  &  61.13  \\
 3  &  61.52 \\
 5  &  \textbf{62.29}  \\
 7  &  61.53 \\
 9  &  61.05 \\
\bottomrule[1.0pt]
\end{tabular}
% \subcaption{Effect of $K_{APA}$.}
\subcaption{}
\label{tab:abl-Kapa}
\end{subtable}

\vspace{0.5em} % 调整行间
\begin{subtable}[t]{0.45\linewidth}
\centering
\begin{tabular}{cc}
\toprule[1.25pt]
S &  Top-1 \\
\midrule[0.75pt]
$0$ & 61.96  \\
$L/2$ & \textbf{62.29}   \\
\bottomrule[1.0pt]
\end{tabular}
% \subcaption{Effect of $S$.}
\subcaption{}
\label{tab:abl-eq7-S}
\end{subtable}
% \hfill
\begin{subtable}[t]{0.45\linewidth}
\centering
\begin{tabular}{cc}
\toprule[1.25pt]
w. $\frac{l}{L}$ &  Top-1 \\
\midrule[0.75pt]
 \checkmark & \textbf{62.29} \\
 $\times$  & 61.32 \\
\bottomrule[1.0pt]
\end{tabular}
% \subcaption{Effect of $\frac{l}{L}$.}
\subcaption{}
\label{tab:abl-eq7-scale}
\end{subtable}
\caption{Effect of varying (a) priors types; (b) $K_{APA}$; (c) $S$; (d) $\frac{l}{L}$.}
\label{tab:ablation_study}
\vspace{-3mm}
\end{table}

% \begin{table}
% \begin{minipage}[t]{0.48\linewidth} % 设置宽度为文本宽度的一半
% \centering
% \begin{tabular}{cc}
% \toprule[1.25pt]
% S &  Top-1 \\
% \midrule[0.75pt]
% $0$ & 61.96  \\
% $L/2$ & 62.29 \\
% % $L$ &  xx \\
% \bottomrule[1.0pt]
% \end{tabular}
% \caption{Effect of $S$.}
% \label{tab:abl-eq7-S}
% \end{minipage}%
% \hfill
% %
% \begin{minipage}[t]{0.48\linewidth} % 调整宽度
% \centering
% \begin{tabular}{cc}
% \toprule[1.25pt]
% w. $\frac{l}{L}$ &  Top-1 \\
% \midrule[0.75pt]
%  \checkmark & 62.29 \\
%  $\times$  & 61.32 \\
% \bottomrule[1.0pt]
% \end{tabular}
% \caption{Effect of $\frac{l}{L}$.}
% \label{tab:abl-eq7-scale}
% \end{minipage}
% \end{table}

% and $\epsilon_1$, and $\epsilon_2$

% \textbf{Analysis on data amount.}
% %
% Tab.\,\ref{tab:abl-imagenumber} presents the effect of using varying numbers of images.
% %
% %
% %
% It can be observed that a progressive improvement in model performance as the image count increases, starting from xx\% accuracy with 32 images and peaking at xx\% with 512 images. For instance, increasing the number of images from 32 to 128 enhances the top-1 accuracy by xxx\% to xx\%. However, beyond 128 images, the performance improvements reach the plateau. Despite using more images can improve the performance, in our main paper, we adopt 32 images in our main experiments to maintain consistency with the previous study~\cite{li2022patch} for a fair comparison. 

\textbf{Analysis of APA loss.}
Here, we conduct the ablation study by considering the $S$ and scale $\frac{l}{L}$ in Eq.\,\ref{eq:apa}. Tab.\,\ref{tab:abl-eq7-S} presents the effect of varying $S$ in Eq.\,\ref{eq:apa}. If applying APA loss to all blocks ($S=0$), the top-1 accuracy decreases to 61.96\%. From Tab.\,\ref{tab:abl-eq7-scale}, it can be seen that absorbing the scale $\frac{l}{L}$ in Eq.\,\ref{eq:apa} presents 0.97\% performance gains.

\section{Limitations}

We further discuss some limitations of the proposed SARDFQ, which will guide future research directions.
First, although SARDFQ shows substantial performance improvement, a performance gap between SARDFQ and real data remains challenging, highlighting the need for a stronger semantics alignment and reinforcement method.
Second, SARDFQ currently lacks a theoretical foundation. Future work could establish a theoretical framework for SARDFQ, particularly in understanding how APA and MSR influence synthetic images in a formalized manner.

% Theoretical exploration would be beneficial for gaining more insight into the DFQ for ViTs.

\section{Conclusion}

In this paper, we investigate the DFQ method for ViTs.
We first identify that synthetic images generated by existing methods suffer from semantic distortion and inadequacy issues, and propose SARDFQ to address these issues. To mitigate semantic distortion, SARDFQ introduces APA, which guides synthetic images to align with randomly generated structural attention patterns. To tackle semantic inadequacy, SARDFQ incorporates MSR. MSR optimizes different regions of synthetic images with unique semantics, thereby enhancing overall semantic richness. Moreover, SARDFQ employs SL, which adopts multiple semantic targets to ensure seamless learning of images augmented by MSR. Extensive experiments on various ViT models and tasks validate the effectiveness of SARDFQ.